%% file: main.tex
\documentclass{article}

\usepackage[final, nonatbib]{neurips_2022}

\usepackage[utf8]{inputenc} 
\usepackage[T1]{fontenc}    
\usepackage{hyperref}       
\usepackage{url}            
\usepackage{booktabs}       
\usepackage{amsfonts}       
\usepackage{nicefrac}       
\usepackage{microtype}      
\usepackage{xcolor}         
\usepackage{graphicx}
\usepackage{subcaption}

\title{Efficient Speech Translation with Pre-trained Models}

\author{Zhaolin Li$^{1,2}$ , Jan Niehues$^{1,2}$\\$^1$Department of Data Science and Knowledge Engineering, Maastricht University, The Netherlands\\$^2$Institute for Anthropomatics and Robotics, Karlsruhe Institute of Technology, Germany\\\{zhaolin.li, jan.niehues\}@kit.edu}

\begin{document}

\maketitle

\begin{abstract}

When building state-of-the-art speech translation models, the need for large computational resources is a significant obstacle due to the large training data size and complex models. The availability of pre-trained models is a promising opportunity to build strong speech translation systems efficiently. In a first step, we investigate efficient strategies to build cascaded and end-to-end speech translation systems based on pre-trained models. Using this strategy, we can train and apply the models on a single GPU. While the end-to-end models show superior translation performance to cascaded ones, the application of this technology has a limitation on the need for additional end-to-end training data. In a second step, we proposed an additional similarity loss to encourage the model to generate similar hidden representations for speech and transcript. Using this technique, we can increase the data efficiency and improve the translation quality by 6 BLEU points in scenarios with limited end-to-end training data.

\end{abstract}

\input{1_introduction}

\input{2_related_work}

\input{3_methodology}

\input{4_experiments_results}

\input{6_conclusion}

\bibliographystyle{IEEEtran}

\bibliography{refs}

\end{document}

%% file: 1_introduction.tex
\section{Introduction}

Building a successful Speech translation (ST) system from scratch is not always possible because of limitations in computation and data resources. Recent works indicate that increasing pre-trained model size still leads to performance improvements on downstream NLP tasks \cite{sanh2020distilbert}. Consequently, the size of the pre-trained model has been getting larger and larger, leading to sometimes impractical to fine-tune the pre-trained models. Collecting the end-to-end data is expensive for finding high-quality data, aligning audio, transcript, and translation, filtering wrong and poor alignment. To address the above challenges, this research focuses on improving computational efficiency and data efficiency with the usage of pre-trained models for speech translation.

Our first contribution is to compare the performances between the cascaded system and the end-to-end system, both using pre-trained models. The end-to-end system has been developed recently and proven to have comparable performance to the cascaded system \cite{Jan2018TheI2, ansari-etal-2020-findings, bentivogli2021cascade}. However, there is no claim about which approach has clear advantages on performance. This work investigates the performance comparison of two systems by directly combining the pre-trained model without architecture modification. Our result shows that the end-to-end system outperforms the cascaded system in terms of fine-tuning efficiency and accuracy. As the second contribution, we propose two fine-tuning strategies to improve computational efficiency. Rather than fine-tuning the entire ST model, we present two effective approaches. Especially, the adapter approach fine-tunes less than one-tenth parameter and achieves comparable performance to the cascaded model. The third contribution is that we present a novel similarity loss to mitigate the data scarcity issue. Unlike the end-to-end data that are challenging to acquire, speech-to-transcript data is more accessible. We develop the similarity loss that measures the difference between latent representations for the audio and the transcript. Our result shows that involving similarity loss improves data efficiency and boosts model performance. 



%% file: 2_related_work.tex
\section{Related work}

The computational limitation is one major obstacle to the end-to-end ST system. \cite{li2021multilingual} presents that fine-tuning the layer normalization and multi-head attention parameters is effective to improve computational efficiency. \cite{le2021lightweight} showed that fine-tuning residual adapt modules that are transplanting between the encoders and decoders is a promising approach. These researches show the possibility of fine-tuning components of pre-trained models and motivate this work to explore other efficient fine-tuning approaches.

The lack of end-to-end training data is another obstacle to the end-to-end ST system. Recent work address this obstacle by leveraging the available data resource using multi-task learning \cite{weiss2017sequencetosequence, anastasopoulos2018tied, berard2018end}, transfer learning \cite{gaido2020endtoend, liu2019end} and generating synthetic data \cite{jia2019leveraging,pino2020self,lam2022sample} techniques. This work extends this idea with a novel loss function to efficiently use the available data.

%% file: 3_methodology.tex
\section{Speech translation using pre-trained models} \label{methodology}

In this work, we propose the cascaded and end-to-end combinations of pre-trained models to build ST systems. A first baseline approach is to combine the two models in a cascaded manner. However, the cascaded approach has several drawbacks, e.g. error propagation and computational complexity. Therefore, we also investigated the possibility of combining the two pre-trained models into one end-to-end speech translation model.

\textbf{Cascaded system}\hspace{0.2cm} In the ASR stage, the module inputs acoustic data and outputs the transcript. In the following MT stage, the transcript gets first segmented into sub-words according to the vocabulary of the translation system. Then the input is fed into the MT module to generate translation.

\begin{figure}[!tbp]
\begin{minipage}[b]{0.45\textwidth}
\includegraphics[scale=0.25]{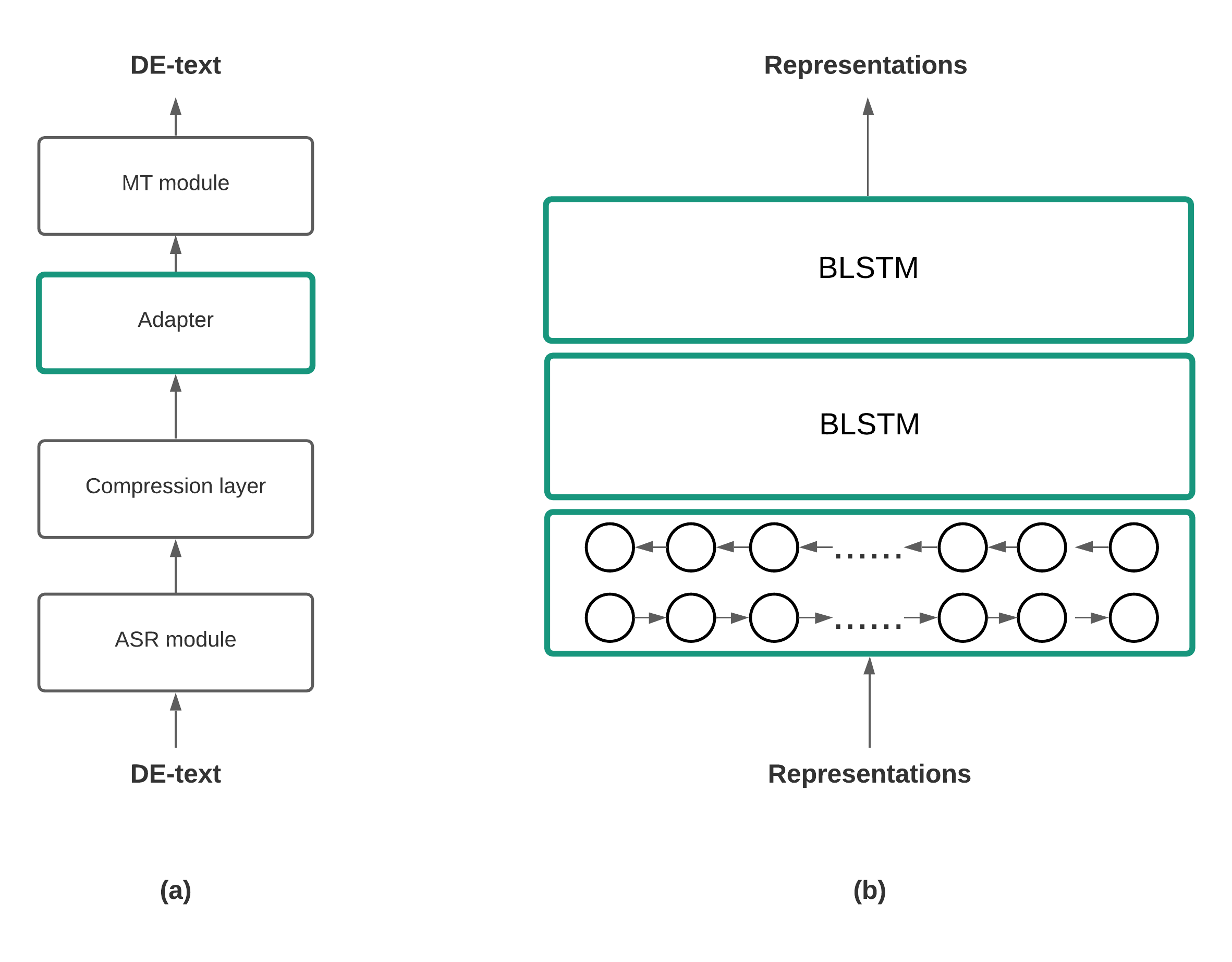}
    \centering
    \caption{Illustration of the adapter. (a) presents the workflow of end-to-end speech translation with the adapter. (b) illustrates the composition of the adapter.}
    \label{fig:adapter}
    \end{minipage}
\hfill
\begin{minipage}[b]{0.5\textwidth}
\includegraphics[scale=0.3]{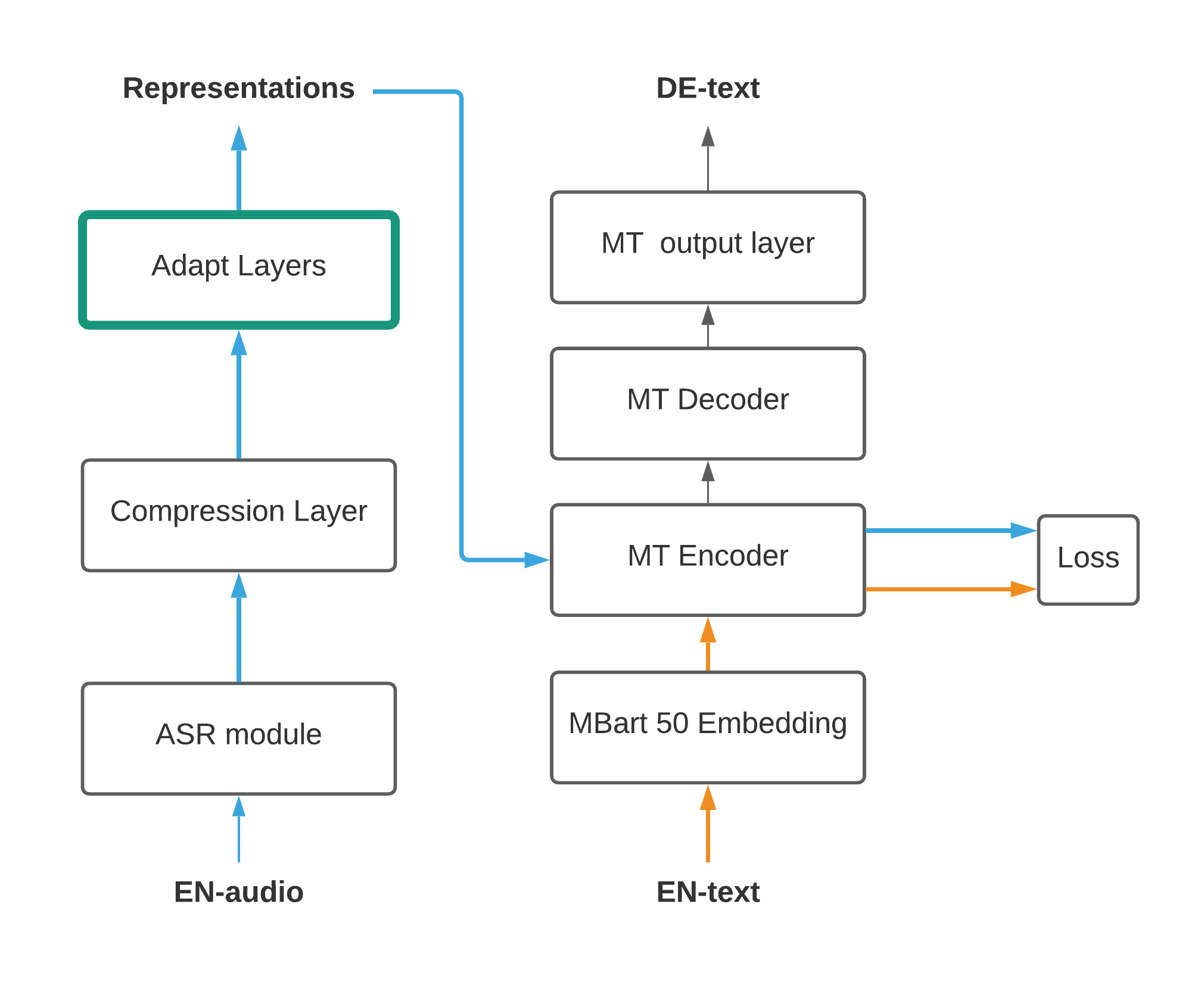}
\centering
\caption{An overview of the similarity loss with an end-to-end system.}
\label{fig:e2e_new_loss}
\end{minipage}
\end{figure}

\textbf{End-to-end system} \hspace{0.2cm} Instead of generating the intermediate transcript, we combine two pre-trained models by feeding the hidden state representation generated from the ASR module to the MT module. The output of the ASR module is character-based. However, the input of the MT module is subword-based. Therefore, the lengths of speech and text sequences are very different for the same segment. The length inconsistency is hard to learn by the ST model, harming model performance. Accordingly, we insert a compression layer between two modules based on the Connectionist Temporal Classification (CTC) algorithm \cite{graves2006connectionist, gaido2021ctcbased}. The layer averages the adjacent speech representations aligned to the same character to compress the redundant and uninformative vectors.

To address the high demand on the memory, we proposed a two-stage training for the end-to-end system to enable training and applying on a single GPU: In the first stage, we fine-tune the pre-trained models on the individual speech recognition or text translation tasks. In this case, all parameters of the model get updated. In the second stage, we jointly train the entire model on the end-to-end task, but only train part of the parameters to improve computational efficiency. Rather than fine-tuning all parameters, we firstly propose only to fine-tune the encoder of the MT module and freeze the rest. The motivation is to solve the discrepancy between the speech representation from the ASR module and the text representation for the decoder. Second, inspired by the adapter investigated in MT \cite{bapna2019simple} and multilingual ST \cite{le2021lightweight}, we propose a simple adapter of three BLSTM layers (Figure \ref{fig:adapter}). The adapter gets inserted between the ASR and MT modules to keep the semantic information of these two modules integrated.

To tackle the lack of the end-to-end data, inspired by \cite{pham2019improving}, we propose a similarity loss function (Figure \ref{fig:e2e_new_loss}). The motivation is that the speech translation model should represent similar hidden state representations for aligned audio and transcript. Consequently, minimizing the similarity loss is proposed to improve speech translation performance. The last hidden states of the MT encoder from EN-audio and DE-text get averaged over time steps to produce the representing vectors. Afterwards, the Mean Squared Error gets calculated between the representing vectors as the loss. As we propose only using the speech-to-transcript training data, the model does not know the target language. Therefore, we implement the target forcing mechanism by generating a target-language-specific embedding with the MT pre-trained embedding layer and prepending the embedding to the speech representations \cite{ha2016toward, johnson2017google, digangi2019onetomany}. Besides, we implement the similarity with the end-to-end system together with the adapter. We only fine-tune the adapter and freeze the rest, to avoid the parameters getting forced to zero to minimize the similarity loss to the optimum zero.

%% file: 4_experiments_results.tex
\section{Experiments \& Results}

The proposed approaches are evaluated on the English-German speech translation of the CoVoST2 \cite{wang2020covost} dataset. In pre-processing, we remove uncompleted data with no transcript or translation, and the double quotes at the beginning and the end of the transcript and translation. Besides, we build a custom vocabulary for ASR tasks which consists of all distinct characters. We use the pre-trained wav2vec2.0 \cite{baevski2020wav2vec} with the large architecture for speech recognition task and a pre-trained MBart50 \cite{liu2020multilingual, tang2020multilingual} for machine translation task.

\begin{table}[ht]
\centering
\caption{Cascaded combinations for the English-German language pair. CoVoST2 indicates the experiment results of the cascaded system from \cite{wang2020covost}. The ASR results are reported in WER, the MT and ST tasks are reported in BLEU. \#params mean how many parameters are fine-tuned.}
\label{tab:cascaded}
\begin{tabular}{ccccc} 
    \toprule 
     & ASR &MT & ST & \#params \\
     \midrule
     Baseline &  29.7 & 32.5 & 16.7 & - \\ 
     Fine-tune wav2vec2.0 & 22.3 & - & 18.4 & 315M \\
     Fine-tune MBart50 & - & 37.3 & 19.5 & 610M\\
     Fine-tune both & - & - & \textbf{21.6} & 925M \\
     Fine-tune MBart50.encoder & - & 36.1 & 18.7 & 152M \\
     CoVoST2 & 21.4 & 29.0 & 18.3 & - \\
    \bottomrule

\end{tabular}
\end{table}

\textbf{Cascaded system} \hspace{0.2cm}In cascaded combination, we explore the efficiency of fine-tuning each component. Firstly we experiment on initializing with parameters of the pre-trained models to provide references. Then, we fine-tune the pre-trained wav2vec2.0 and MBart50 models with speech-to-transcript and transcript-to-translation training data, respectively. We experiment with different combinations of the pre-trained and fine-tuned parameters to explore efficiency. 
As Table \ref{tab:cascaded} shows, fine-tuning both modules leads to the best improvements of 4.9 BLEU points compared with no fine-tuning and 3.3 BLEU points improvement on the CoVoST2 Cascaded. Besides, we find that fine-tuning the encoder of the MT model slightly improves performance. With 25\% parameters, fine-tuning the encoder achieves 41\% improvements of fine-tuning the entire MBart50.

\begin{table}[ht]
\centering
\caption{End-to-end combination experiments results. ASR and MT mean the parameter initialization of the end-to-end model components. PT means pre-trained, FT means fine-tuned. Training strategy indicates which component is fine-tuned. \textit{No} means the experiment has no fine-tuning. CoVoST2 indicates the experiment results of the end-to-end system from \cite{wang2020covost}.}
\label{tab:e2e}
\begin{tabular}{cccccc} 
    \toprule 
     & ASR & MT & Training  & \verb|#|params & BLEU \\
     \midrule
    E1 & PT & PT & No & - & 0 \\ 
    E2 & FT & PT & No & - & 0.5 \\ 
    E3 & FT & FT & No & - & 0.1 \\ 
    E4 & FT & PT & MT.encoder & 152M & \textbf{22.8} \\ 
    E5 & PT & PT & MT.encoder & 152M & 0.4 \\ 
    E6 & FT & FT & MT.encoder & 152M &22.0 \\ 
    E7 & FT & PT & Adapt layers & 67M & \textbf{20.9} \\
    CoVoST2 & - & - & - & - & 16.3 \\
    \bottomrule

\end{tabular}
\end{table}

\textbf{End-to-end system} \hspace{.2cm}In the end-to-end combination, we apply a two-stage training scheme (Section \ref{methodology}). The first stage is fine-tuning the pre-trained models, and the second stage is fine-tuning the end-to-end model with the end-to-end data. In light of building the cascaded model directly using the ASR and MT modules, we first experiment with the initialization on the end-to-end system. As Table \ref{tab:e2e} shows, the end-to-end model does not work without the end-to-end training data according to experiments 1, 2 and 3. Next, we experiment with the end-to-end training data on different fine-tuning modules in the first stage and fine-tuning the MT encoder or adapter in the second stage to address computation efficiency. As shown in experiments 4, 5 and 6, the two-stage approach that trains the ASR component independently in the first stage and the MT encoder using the end-to-end data in the second stage is promising to build end-to-end ST systems. With this configuration (E4), we can achieve a better translation quality than the cascaded system. Furthermore, the end-to-end combination achieves 4.5 BLEU points improvements compared with the CoVoST2 E2E.

A second approach is integrating an additional adapt layer (E7), where we only need to train 67M instead of 150M parameters. The performance is 2 BLEU points worse. However, the pre-trained MT model is not changed and therefore can, for example, still be used for text translation in parallel.

\textbf{Similarity loss} \hspace{.2cm} In a second series of experiments, we evaluated the data efficiency of the end-to-end model with respect to end-to-end training data. Therefore, we investigated the effect of the similarity loss on the best-performing system using the adapt layer (E2E 7). To use the same hyperparameters as the previous experiments in model training, we scale up the similarity loss value by 100 to make it at the same scale as the original experiments. We evaluate model performance on different portions of the training data to evaluate the data efficiency.

\begin{table}
\centering
\caption{Experiment results on the similarity loss. For similarity loss, \textit{without} means training only with the similarity loss; the following portion means continuing from the \textit{without}, training with that portion of data with the original loss. The result is reported in the BLEU score.}
\begin{tabular}{cccccc}
\hline
    Experiment & Without & 10\% & 15\% & 20\% & All  \\
    \hline
    Original loss & - & 0.3 & 2.0 & 11.8 & 20.9 \\ 
    Similarity loss & 0 & 1.0 & 10.7 & 17.8 & 21.6 \\
    \hline
\end{tabular}
\label{tab:loss}
\end{table}

In a first experiment, we evaluated the model on the zero-shot condition, where no end-to-end training data was available. As Table \ref{tab:loss} shows, this approach fails to enable speech translation. We observe that the translation is in the source language, although with the target forcing mechanism. Therefore we expect that involving a few end-to-end data would solve the issue. Continue from the model trained with the similarity loss, we experiment on training with the original loss using different amounts of data. We observe that training with 10\% training data enables the model to translate into the correct target language but poorly. With 20\% data, adding similarity loss improves 51\% compared with that without the loss. The evaluation score reaches 17.8 BLEU points, achieving 85\% performance of the best end-to-end model. Besides, compared with the learning curve of the original loss, adding the similarity loss enables the model to fulfil speech translation tasks with less training data. The advantages of adding the similarity loss demonstrate a promising approach to improving data efficiency. In addition, we observe that with all training data, training the model with the similarity loss gains 0.7 BLEU point improvement. Therefore, we conclude that involving the similarity loss increases data efficiency and benefits to improving model performance.

%% file: 6_conclusion.tex
\section{Conclusion}
In this work, we proposed using pre-trained speech recognition and text translation models to build a state-of-the-art speech translation system with limited resources. While a cascaded combination directly achieves relatively good performance, we develop several techniques to enable the end-to-end system to use these models, and handle different word representations used in the pre-trained models. Secondly, we propose two training strategies that allow the training and inference on a single GPU. Finally, we present an additional training loss to reduce the need for end-to-end training data. Using all these techniques, the proposed end-to-end model can outperform the cascaded model.